\documentclass[11pt]{article}

\usepackage{acl}

\usepackage{times}
\usepackage{latexsym}
\usepackage[T1]{fontenc}
\usepackage[utf8]{inputenc}
\usepackage{microtype}
\usepackage{inconsolata}
\usepackage{graphicx}
\usepackage{booktabs}
\usepackage{amsmath}
\usepackage{amssymb}
\usepackage[most]{tcolorbox}

\hypersetup{
  pdftitle={Auditing Evidence Use in Medical LLM Diagnosis},
  pdfauthor={Junchi Liao, Jiawen Deng, and Fuji Ren}
}

\definecolor{PromptBlueBack}{HTML}{EFF6FF}
\definecolor{PromptBlueFrame}{HTML}{2563EB}
\definecolor{PromptGreenBack}{HTML}{ECFDF5}
\definecolor{PromptGreenFrame}{HTML}{059669}
\definecolor{PromptOrangeBack}{HTML}{FFF7ED}
\definecolor{PromptOrangeFrame}{HTML}{EA580C}
\definecolor{PromptPurpleBack}{HTML}{F5F3FF}
\definecolor{PromptPurpleFrame}{HTML}{7C3AED}

\newtcolorbox{diagnosisprompt}[1]{
  breakable,
  colback=PromptBlueBack,
  colframe=PromptBlueFrame,
  boxrule=0.5pt,
  arc=2pt,
  left=4pt,
  right=4pt,
  top=4pt,
  bottom=4pt,
  fonttitle=\bfseries\small,
  title={#1}
}
\newtcolorbox{subsetprompt}[1]{
  breakable,
  colback=PromptGreenBack,
  colframe=PromptGreenFrame,
  boxrule=0.5pt,
  arc=2pt,
  left=4pt,
  right=4pt,
  top=4pt,
  bottom=4pt,
  fonttitle=\bfseries\small,
  title={#1}
}
\newtcolorbox{neutralprompt}[1]{
  breakable,
  colback=PromptOrangeBack,
  colframe=PromptOrangeFrame,
  boxrule=0.5pt,
  arc=2pt,
  left=4pt,
  right=4pt,
  top=4pt,
  bottom=4pt,
  fonttitle=\bfseries\small,
  title={#1}
}
\newtcolorbox{reviewprompt}[1]{
  breakable,
  colback=PromptPurpleBack,
  colframe=PromptPurpleFrame,
  boxrule=0.5pt,
  arc=2pt,
  left=4pt,
  right=4pt,
  top=4pt,
  bottom=4pt,
  fonttitle=\bfseries\small,
  title={#1}
}

\title{Auditing Evidence Use in Medical LLM Diagnosis}

\author{%
  Junchi Liao \quad Jiawen Deng \quad Fuji Ren\thanks{Corresponding author.} \\
  University of Electronic Science and Technology of China \\
  Chengdu, China
}

\begin{document}
\maketitle

\begin{abstract}
Medical LLMs are often evaluated by whether they select the correct diagnosis, but diagnostic accuracy alone does not show whether the model used the case evidence appropriately. We present a behavioral audit of evidence use in medical diagnosis. For each case, we decompose patient information into evidence units, score candidate diagnoses under controlled evidence subsets, and mine low-order interactions in diagnostic margins. Because medical evidence is diagnosis-relative, the audit separates interaction discovery from failure assignment: large or negative interactions can reflect plausible differential diagnosis, while suspicious interactions require robustness checks and clinical review. We evaluate five open-weight LLMs on DDXPlus, CupCase, and MedCase. Across datasets, faithful support and differential conflict or cancellation account for most interaction strength, showing that many evidence interactions are clinically plausible rather than failures. In a DDXPlus-focused blinded five-reviewer 130-item enriched review sample, invalid or shortcut-like cases concentrate in negated or absent findings and clinically local evidence. These results show that accuracy can hide candidate evidence-use failures and motivate role-aware audits for medical LLM evaluation.
\end{abstract}

\section{Introduction}

Large language models (LLMs) are increasingly evaluated as diagnostic assistants in medicine, with foundation-model work suggesting broad potential for generalist medical AI \citep{moor2023foundation,thirunavukarasu2023large,clusmann2023future}. Most evaluations ask whether the final diagnosis or answer is correct, often through medical exams or case-based question answering benchmarks \citep{kung2023performance,nori2023capabilities,singhal2023large,singhal2025toward}. This is a necessary signal, but it is not sufficient for high-stakes clinical use. A model can select the right diagnosis while giving too much weight to evidence that a clinician would consider weak, irrelevant, or directionally misleading.

The gap is especially important for diagnosis because clinical evidence is not a flat list of features. Symptoms, signs, risk factors, negative findings, and history items play different roles in a differential diagnosis, a structure reflected in diagnostic datasets that expose patient findings and candidate differentials \citep{fansi2022ddxplus}. The same finding can support one disease, support a competing disease, or rule against another; an explicitly absent finding may be informative for one diagnosis while irrelevant to another. Accuracy collapses these distinctions into a single final label, a known limitation of aggregate benchmark reporting in broader LLM evaluation \citep{hendrycks2020measuring,srivastava2023beyond,liang2022holistic}.

We study this problem as an evidence-use auditing task, following behavioral and counterfactual evaluation work that tests model behavior under controlled input changes \citep{ribeiro2020beyond,gardner2020evaluating}. Given a clinical case, we split the patient information into evidence units and present the model with controlled combinations of those units. The model chooses among candidate diagnoses, and we score the diagnostic margin for each candidate from the same multiple-choice prompt. We then mine low-order evidence interactions: combinations of evidence units whose joint presence changes the diagnostic margin beyond their individual effects.

This audit is behavioral rather than introspective. We do not claim to recover the model's hidden reasoning process or to validate generated rationales, since rationale and chain-of-thought faithfulness can be fragile \citep{jacovi2020towards,turpin2023language,lanham2023measuring}. Instead, we ask a narrower question: when clinical evidence is made available or unavailable under controlled interventions, does the model's observable diagnostic preference change in a clinically coherent way?

This distinction matters because medical interactions are not errors by default. A negative interaction can reflect legitimate differential diagnosis: evidence for pneumonia may lower the margin for asthma, and overlapping asthma findings may be redundant rather than additive. We therefore separate interaction discovery from failure assignment. Interaction mining proposes candidate mechanisms; diagnosis-relative evidence roles, robustness checks, and clinical validation determine which candidates support stronger failure claims.

This distinguishes the audit from three adjacent evaluation styles: diagnostic benchmarks that stop at the full-case answer, attribution or rationale studies that evaluate token salience or explanation faithfulness, and counterfactual tests that treat a feature as globally unwanted or irrelevant \citep{lundberg2017unified,sundararajan2020shapley,deyoung2020eraser}. Here, evidence is diagnosis-relative: the same field can support one diagnosis, support a competitor, or be clinically incoherent for another. The contribution is therefore the role-aware audit protocol around interaction mining, not the interaction score alone.

Our contributions are:
\begin{itemize}
\item We propose a behavioral audit for medical diagnosis that measures how LLM diagnostic margins change under controlled evidence interventions.
\item We introduce a diagnosis-relative interpretation framework that separates plausible differential-diagnosis interactions from candidate evidence-use failures.
\item We evaluate five open LLMs on three diagnostic datasets, showing that broad faithful-support and conflict/cancellation patterns persist across case styles; on DDXPlus, clinically adjudicated invalid mechanisms concentrate in evidence polarity and clinical specificity.
\end{itemize}

\section{Method}

\begin{figure*}[t]
\centering
\includegraphics[width=0.98\textwidth]{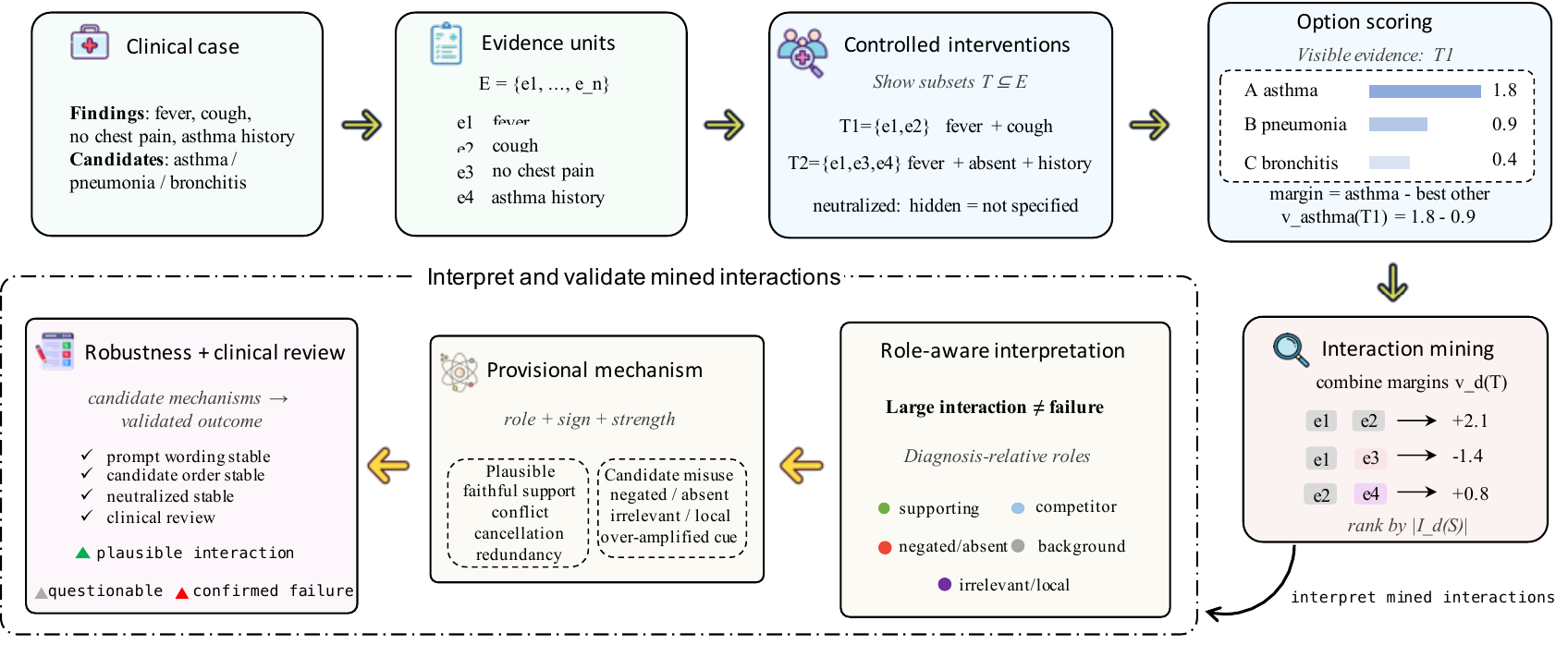}
\caption{Overview of the evidence-use audit. Cases are decomposed into evidence units, scored under controlled interventions, and mined for margin interactions; these interactions are interpreted with diagnosis-relative roles and validated before failure claims are assigned.}
\label{fig:method-overview}
\end{figure*}

The audit turns a clinical case into controlled observations of model preference: it scores diagnoses under evidence interventions, assigns diagnosis-relative evidence roles, mines low-order interactions, and maps stable high-strength interactions to clinical mechanisms. It surfaces candidate evidence-use patterns first; failure labels are assigned only after robustness checks and clinical interpretation. Figure~\ref{fig:method-overview} summarizes the full pipeline.

\subsection{Diagnostic Scoring with Evidence Interventions}

Given a clinical case, we represent patient information as evidence units $E=\{e_1,\ldots,e_n\}$ and candidate diagnoses $C=\{c_1,\ldots,c_m\}$. Units include symptoms, signs, risk or history items, laboratory or imaging statements, and explicitly absent findings. For datasets with a gold diagnosis, the target is $y\in C$, matching the multiple-choice format used by many medical QA and diagnostic benchmarks \citep{jin2019pubmedqa,jin2021disease,pal2022medmcqa}.

Each prompt shows the visible evidence units and the same candidate list, and asks the model to return only the option letter. Rather than using only the generated answer, we score every candidate option from the same prompt: in the main experiments, $s_T(c_j)$ is the next-token log probability of the option letter for diagnosis $c_j$. The prediction under subset $T$ is $\arg\max_{c_j\in C}s_T(c_j)$.

Our audit uses a scalar diagnostic preference score rather than the raw prediction alone. For any audited diagnosis $d\in C$, we define its margin under subset $T$ as
\begin{equation}
v_d(T) = s_T(d) - \max_{c_j\in C, c_j\neq d} s_T(c_j).
\end{equation}
Larger margins indicate stronger preference for $d$ over its closest competitor. We use $d=y$ for gold-diagnosis analysis and the same definition for competitors. Margins are within-prompt preference differences, not calibrated diagnostic probabilities; option-letter priors and position effects can still affect scores, so stronger failure claims require candidate-order robustness.

Evidence interventions vary $T$ over the selected audited units. In the main setting, selected audited units outside $T$ are hidden from the prompt, with dataset-specific fixed context handled as described in Appendix~\ref{app:robustness-details}. In sensitivity analyses, hidden units are instead replaced with a ``not specified'' statement, and margins and interactions are recomputed under that perturbation. This matters for negated evidence: removing ``no chest pain'' makes the finding unavailable; it does not assert that chest pain is present.

\subsection{Diagnosis-Relative Evidence Roles}

Medical evidence is diagnosis-relative: a symptom may support one diagnosis, support a competitor, or exclude another candidate. We therefore assign evidence roles relative to the audited diagnosis rather than treating a unit as globally relevant or irrelevant.

For an evidence unit $e_i$ and audited diagnosis $d$, role families are: \emph{supporting evidence}, which makes $d$ more plausible; \emph{competitor-supporting evidence}, which supports an alternative and can lower $d$'s margin; \emph{excluding or negated evidence}, which reports absence or rules against $d$; \emph{background or risk evidence}, which changes prior plausibility; and \emph{likely irrelevant or clinically local evidence}, which should have little direct effect without a clinical bridge.

These roles are directional expectations, not hard rules: absent symptoms can be weakly informative, and background history can matter in some differentials. Role annotation guides interpretation of interaction sign and strength but does not itself determine validity. Appendix Table~\ref{tab:appendix-roles} gives the full role summary.

\subsection{Interaction Mining}

Single-evidence effects are often insufficient for diagnosis because findings can become informative only in combination. We therefore mine interactions over evidence units: the part of a diagnostic-margin change attributable to a set of units jointly, beyond lower-order effects.

Given the value function $v_d(T)$ over evidence subsets, we compute the interaction effect for a non-empty set $S$ with respect to diagnosis $d$ as
\begin{equation}
I_d(S) = \sum_{T\subseteq S}(-1)^{|S|-|T|}v_d(T).
\end{equation}
For a singleton $S=\{e_i\}$, this is the effect of one unit relative to the empty-evidence baseline. For larger $S$, it measures whether units jointly support or suppress diagnosis $d$ beyond their individual effects. Positive values increase the margin for $d$; negative values reduce it or favor a competitor.

We refer to $|S|$ as interaction order and $|I_d(S)|$ as strength. Low-order interactions are the main audit targets because they are interpretable and cheaper to estimate, following attribution and interaction work on joint feature effects \citep{grabisch1999axiomatic,lundberg2017unified,sundararajan2020shapley,janizek2021explaining,tsang2017detecting}. We audit a fixed selected set of units per case, rank high-strength interactions, and group them by diagnosis-relative roles and signs.

\subsection{From Interactions to Audit Labels}

Because large interactions can reflect valid differential diagnosis or redundancy, automatic labels are provisional. Rules use the interaction sign, strength, audited diagnosis, candidate set, and evidence roles, and are applied in a deterministic priority order. The pipeline first checks candidate-misuse conditions: (i) positive interactions containing explicitly negated or excluding evidence, then (ii) interactions containing clinically local evidence with no role-supported path to the audited diagnosis. These checks take precedence over later faithful or conflict labels when an interaction satisfies multiple conditions.

If neither candidate-misuse condition applies, the remaining interactions fall through to mechanism labels. Competitor-supporting evidence is labeled differential conflict/cancellation when the audited margin is reduced or target and competitor evidence are mixed; suppressive interactions among target-supporting units are target redundancy; positive target-supporting interactions are faithful target evidence; antecedent, risk, exposure, or prior-condition roles are background/history influence. Remaining low-strength or ambiguous interactions are weak or redundant. Appendix Table~\ref{tab:appendix-labels} gives label definitions.

A smaller subset is marked as candidate failure mechanisms: \emph{negated or absent evidence misuse}, where absent or excluding findings increase support unexpectedly, and \emph{irrelevant evidence misuse}, where clinically unrelated or overly local fields strongly affect the margin. These are suspicious interactions, not final judgments. We reserve strong failure claims for interactions that pass stability checks across prompt wording, candidate order, and perturbation semantics and are judged invalid; questionable cases are reported separately.

\section{Experiments}

\subsection{Experimental Setup}

We evaluate five open-weight instruction-tuned LLMs: Qwen2.5-7B-Instruct, OpenBioLLM-8B, MedGemma-4B, BioMistral-7B, and Llama3-Med42-8B \citep{hui2024qwen2,OpenBioLLMs,sellergren2026medgemmatechnicalreport,labrak2024biomistral,christophe2024med42}. DDXPlus provides structured symptom and history fields for fine-grained role analysis \citep{fansi2022ddxplus}; CupCase and MedCase use case-report style descriptions and test whether broad interaction patterns persist beyond structured DDXPlus \citep{perets2025cupcase,wu2025medcasereasoning}.

\begin{table}[h]
\centering
\small
\begin{tabular}{lrrrr}
\toprule
Dataset & Cases & Units & Cand. & Interv. \\
\midrule
DDXPlus & 500 & 8 & 2--20 & 255 \\
CupCase & 200 & 6 & 3--4 & 63 \\
MedCase & 200 & 6 & 3--8 & 63 \\
\bottomrule
\end{tabular}
\caption{Evaluation data. ``Units'' denotes the number of selected evidence units audited per case. ``Cand.'' is the range of candidate diagnoses, and ``Interv.'' is the number of subset interventions per case, excluding the full-evidence prompt.}
\label{tab:datasets}
\end{table}

For each case, we score candidate diagnoses under evidence interventions, compute interactions over selected units, and report accuracy, interaction mechanisms, clinical validation, targeted counterfactuals, and robustness checks.

DDXPlus supports structured role labels, so fine-grained failure labels and clinical-review claims are mainly analyzed there. CupCase and MedCase serve as coarser external checks for faithful-support versus conflict/cancellation structure, not fine-grained failure prevalence.
Appendix~\ref{app:experimental-setting} provides dataset construction, evidence-unit selection, candidate construction, and model-scoring details; Appendix~\ref{app:prompts} gives the prompt templates.

\subsection{Main Results}

\paragraph{Finding 1: Accuracy is not evidence-use fidelity.}
Table~\ref{tab:accuracy} gives the standard diagnostic baseline. OpenBioLLM has the strongest average accuracy, and Med42 is close on DDXPlus; CupCase is easier for most models, while MedCase is harder, with the best model reaching 55.0\%.

Accuracy is useful but does not identify the evidence-use mechanism behind a prediction: correct answers can arise from faithful evidence, plausible cancellation among competitors, or irrelevant/absent findings that shift the margin in the right direction.

Comparing Table~\ref{tab:accuracy} with Appendix Table~\ref{tab:appendix-model-mechanisms}, the five evaluated models do not have the same ordering by DDXPlus accuracy as by faithful target-evidence strength or conflict/cancellation strength. We treat this as a qualitative observation rather than a correlation estimate; the practical point is that accuracy and evidence-use structure are not interchangeable summaries.

\begin{table}[h]
\centering
\small
\setlength{\tabcolsep}{3.1pt}
\begin{tabular}{lrrrrr}
\toprule
Dataset & Qwen & OpenBio & MedG. & BioMis. & Med42 \\
\midrule
DDXPlus & 65.8 & 74.0 & 64.0 & 54.6 & 73.4 \\
CupCase & 87.0 & 87.5 & 70.5 & 80.0 & 77.0 \\
MedCase & 40.5 & 55.0 & 43.5 & 44.5 & 49.0 \\
\midrule
Avg. & 64.4 & 72.2 & 59.3 & 59.7 & 66.5 \\
\bottomrule
\end{tabular}
\caption{Full-evidence diagnostic accuracy (\%) across models and datasets. MedG. denotes MedGemma and BioMis. denotes BioMistral.}
\label{tab:accuracy}
\end{table}

\begin{figure}[h]
\centering
\includegraphics[width=\columnwidth]{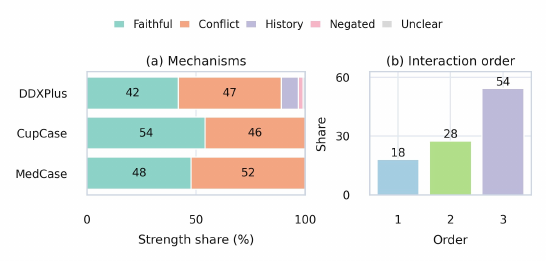}
\caption{Evidence-use mechanisms. (a) Share of interaction strength by clinical mechanism across datasets. (b) Distribution of interaction strength by interaction order on DDXPlus.}
\label{fig:evidence-mechanisms}
\end{figure}

\paragraph{Finding 2: Medical interactions are common, but many are legitimate.}
Figure~\ref{fig:evidence-mechanisms} shows that diagnostic margins are not driven only by isolated findings. Differential conflict/cancellation accounts for 47.1\% of interaction strength on DDXPlus, 45.8\% on CupCase, and 52.2\% on MedCase, while faithful or plausible support accounts for much of the remainder. Most salient DDXPlus strength comes from low-order combinations, especially third-order interactions, which are richer than singleton attribution but still inspectable.

Table~\ref{tab:order-baseline} compares lower-order perturbation baselines with the order-3 audit on DDXPlus. Singleton or leave-one-out scoring uses only 9 subsets per case but keeps 18.0\% of top-interaction strength, retrieves 6/300 raw suspicious interactions, and retrieves none of the 11 adjudicated invalid or shortcut-like review items. Pairwise interactions recover more signal but only 4/11 invalid or shortcut-like items. Because the 11 reviewed invalid or shortcut-like items were drawn from the order-3 output, the order-3 row is a reference set rather than an independent sufficiency result; the meaningful comparison is the coverage gap of the lower-order baselines.

\begin{table}[h]
\centering
\small
\setlength{\tabcolsep}{2.4pt}
\begin{tabular}{lrrrr}
\toprule
Scope & Budget & Str. kept & Raw susp. & Invalid \\
\midrule
Singleton & 9 (3.5\%) & 18.0\% & 6/300 & 0/11 \\
Pairwise & 37 (14.5\%) & 45.6\% & 80/300 & 4/11 \\
Order 3 & 93 (36.3\%) & 100.0\% & 300/300 & 11/11 \\
\bottomrule
\end{tabular}
\caption{Order-ablation baseline on DDXPlus 500x8. Budget is scored subsets per case, with 256 full subset combinations for eight audited units. Strength is cumulative top-interaction absolute strength retained within the order-3 audit. Raw susp. is the DDXPlus 300-item automatic suspicious queue, and Invalid is the adjudicated invalid or shortcut-like subset of the enriched 130-item clinical review sample.}
\label{tab:order-baseline}
\end{table}

The key point is that interaction strength has different clinical meanings. Table~\ref{tab:ddxplus-mechanisms} shows this on DDXPlus: conflict/cancellation and faithful target evidence dominate total strength, while background/history evidence is less frequent but larger on average. This pattern is not driven by one model: faithful/plausible support plus conflict/cancellation accounts for 86.6--89.6\% of DDXPlus strength across the five models, while negated/absent misuse remains 1.2--3.0\%. Appendix Tables~\ref{tab:appendix-model-mechanisms}, \ref{tab:appendix-dataset-mechanisms}, and~\ref{tab:appendix-order} report model-wise, dataset-wise, and interaction-order breakdowns.

\begin{table}[h]
\centering
\small
\setlength{\tabcolsep}{3.2pt}
\begin{tabular}{lrrr}
\toprule
Mechanism & Count & Strength & Mean \\
\midrule
Conflict/cancel. & 40.0 & 42.7 & 2.51 \\
Faithful target & 37.7 & 36.6 & 2.29 \\
Background/history & 4.1 & 7.9 & 4.55 \\
Weak/redundant & 7.1 & 5.3 & 1.76 \\
Target redundancy & 3.0 & 4.4 & 3.49 \\
Negated misuse & 2.5 & 2.2 & 2.12 \\
\bottomrule
\end{tabular}
\caption{Fine-grained DDXPlus mechanism distribution. Count and strength are percentages; Mean is the average absolute interaction effect.}
\label{tab:ddxplus-mechanisms}
\end{table}

\subsection{Clinical Validation}

\paragraph{Finding 3: DDXPlus invalid mechanisms concentrate in evidence polarity and specificity.}
Automatic labels are screening signals, so we clinically reviewed 130 high-strength DDXPlus interactions sampled across mechanism buckets. Five senior PhD-trained medical reviewers independently annotated all items under a blinded protocol; source model and provisional bucket were hidden. Before adjudication, all five reviewers agreed on 77\% of items, with high inter-rater agreement (Fleiss' $\kappa=0.82$).

\begin{figure}[h]
\centering
\includegraphics[width=\columnwidth]{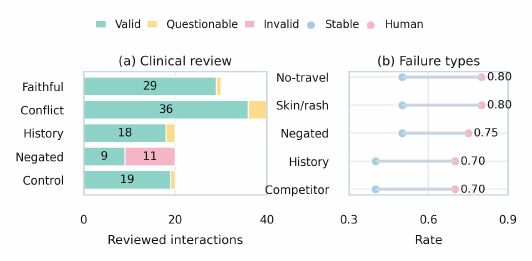}
\caption{Clinical validation. (a) Adjudicated review outcomes for sampled interactions by mechanism. (b) Stability and human-validation summaries for candidate failure types.}
\label{fig:clinical-validation}
\end{figure}

Figure~\ref{fig:clinical-validation} and Table~\ref{tab:clinical-review} separate discovery from error confirmation. After adjudication, 111/130 interactions are valid, 8 questionable, and 11 invalid or shortcut-like. Faithful target evidence is almost always valid, background/history is often legitimate, and conflict/cancellation has no adjudicated invalid cases in this DDXPlus sample. Appendix Tables~\ref{tab:appendix-review-sample}, \ref{tab:appendix-review-protocol}, and~\ref{tab:appendix-review-confidence} give sampling, protocol, and confidence details.

Invalid mechanisms concentrate in the DDXPlus enriched review sample's negated/absent bucket. The 11 invalid cases include no recent travel increasing support for allergic sinusitis, asthma exacerbation, or Guillain-Barre syndrome, and skin/rash or no-peeling fields supporting unrelated diagnoses such as inguinal hernia or influenza. Thus, the substantive DDXPlus finding is a stable cluster around evidence polarity and overly broad transfer of clinically local fields.

\begin{table}[h]
\centering
\small
\setlength{\tabcolsep}{4pt}
\begin{tabular}{lrrr}
\toprule
Bucket & Valid & Quest. & Invalid \\
\midrule
Faithful target & 29 & 1 & 0 \\
Conflict/cancel. & 36 & 4 & 0 \\
Background/history & 18 & 2 & 0 \\
Negated misuse & 9 & 0 & 11 \\
Control/unclear & 19 & 1 & 0 \\
\midrule
Total & 111 & 8 & 11 \\
\bottomrule
\end{tabular}
\caption{Adjudicated clinical review counts for 130 sampled interactions. Quest. denotes clinically questionable cases.}
\label{tab:clinical-review}
\end{table}

\subsection{Targeted Counterfactual Validation}

\paragraph{Finding 4: Counterfactuals separate local artifacts from global reliance.}
Clinical review judges plausibility, but targeted counterfactuals test whether suspicious evidence only appears in the decomposition or changes the model's diagnostic preference. For representative audit-surfaced cases, we remove or rewrite the suspicious evidence family while keeping the rest of the case fixed and rescore the same candidates.

\begin{table}[h]
\centering
\small
\setlength{\tabcolsep}{2.6pt}
\begin{tabular}{llrl}
\toprule
Pattern & Counterfactual & $\Delta$ margin & Rank \\
\midrule
MG name cue & remove MG history & $-9.50$ & $1\!\to\!1$ \\
MG name cue & remove MG history & $-6.13$ & $1\!\to\!2$ \\
Hernia skin & remove skin fields & $+2.25$ & $1\!\to\!1$ \\
Influenza skin & remove skin fields & $+3.75$ & $2\!\to\!1$ \\
SLE skin control & remove skin fields & $-7.50$ & $1\!\to\!2$ \\
\bottomrule
\end{tabular}
\caption{Representative targeted counterfactuals. MG denotes myasthenia gravis. The table reports target-margin changes after removing the indicated evidence family; rows are representative model-case pairs from the targeted counterfactual validation.}
\label{tab:counterfactual}
\end{table}

Table~\ref{tab:counterfactual} sharpens the audit interpretation. Exact disease-name family history strongly amplifies confidence for myasthenia gravis: removing it reduces the target margin by $9.50$ for Qwen while preserving the top diagnosis, and by $6.13$ for BioMistral while flipping the top diagnosis. This supports a confidence-amplification claim, not a claim that the cue is always the sole reason for the diagnosis.

Skin/rash patterns illustrate the need for this check. For inguinal hernia, skin fields form clinically implausible local interactions, but removing them does not break the final prediction; the claim is local incoherence, not global reliance. For influenza, removal often improves target margin or rank, suggesting competitor confusion. The SLE control gives the boundary case: skin evidence can be clinically meaningful, and removing it can hurt the target.

An expanded 86-record stratified validation supports this narrower interpretation. On clean-correct model-case pairs, high no-travel cases have only modestly higher strong-change rates than low no-travel controls (11.6\% vs. 5.6\%), so no-travel remains a secondary polarity example. Skin/rash strata are stronger: high skin-hernia and high skin-other reach 29.3\% and 32.1\% strong-change rates, with larger competitor-suppression rates (44.3\% and 46.2\%) than direct target-score decreases (5.2\% and 3.8\%). The plausible-skin control is also sensitive (45.0\%), matching the clinical boundary case rather than contradicting it. Appendix Table~\ref{tab:appendix-stratified-counterfactual} gives the full breakdown.

\subsection{Robustness and Sensitivity}

\paragraph{Finding 5: Stability filtering turns broad suspicious patterns into high-confidence candidates.}
Raw automatic suspicious interactions are too broad to report as failures. The check targets three artifact channels: instruction framing, option-position bias, and whether deletion itself creates an artificial signal. We therefore test stability to prompt wording, candidate order, and perturbation semantics: prompt/order variants on 100 DDXPlus cases and deletion versus neutralization on 200 cases.

\begin{figure}[h]
\centering
\includegraphics[width=\columnwidth]{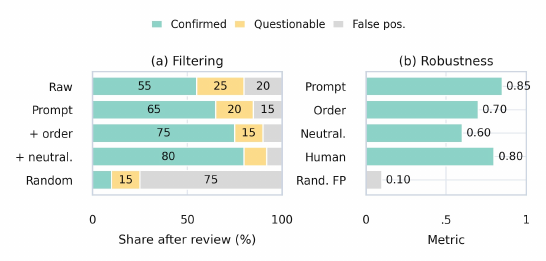}
\caption{Robustness and filtering. (a) Stability filtering improves the precision of candidate evidence-use failures. (b) Summary of prompt, order, perturbation, human-validation, and random-control checks.}
\label{fig:robustness}
\end{figure}

Figure~\ref{fig:robustness} and Table~\ref{tab:filtering} show the filtering effect on an enriched human-adjudicated candidate-failure queue. Stability filtering reduces the queue from 300 to 120 interactions; within this queue, adjudicated precision rises from 0.55 to 0.80 and false-positive rate falls from 0.20 to 0.08. Random controls have much lower precision and higher false positives. These are descriptive queue-level summaries, not population estimates.

Tables~\ref{tab:prompt-order} and~\ref{tab:neutralize} provide sensitivity checks. Prompt and option-order changes affect behavior, as expected for option-scored LLMs, but suspicious-strength ratios stay in the same order of magnitude across prompt families and option orders. Neutralization reduces the suspicious signal for every model, sharply for OpenBio and BioMistral, but does not eliminate it. These results support a cautious interpretation: not every mined interaction is a failure, but the stable subset is not explained away by prompt wording, option order, or deletion artifacts.

Because evidence-unit selection can also affect which local interactions are available to mine, we additionally vary the DDXPlus unit-selection strategy. Under target-salient and random-balanced selectors, faithful/plausible support plus conflict/cancellation accounts for 94.0\% and 94.0\% of strength, compared with 89.0\% under the main balanced selector. Suspicious strength remains small (2.4--2.8\%), and negated/absent misuse remains 2.1--2.7\%. Top-interaction overlap drops to about 0.5 because the audited units change, so this is mechanism-level rather than exact-interaction stability. Appendix Table~\ref{tab:unit-selection-sensitivity} gives the full selection-sensitivity results.

\begin{table}[h]
\centering
\small
\setlength{\tabcolsep}{3.2pt}
\begin{tabular}{lrrrr}
\toprule
Candidate set & $n$ & Prec. & Quest. & FP \\
\midrule
Raw automatic & 300 & 0.55 & 0.25 & 0.20 \\
Prompt stable & 220 & 0.65 & 0.20 & 0.15 \\
+ order stable & 180 & 0.75 & 0.15 & 0.10 \\
+ neutralize & 120 & 0.80 & 0.12 & 0.08 \\
Random control & 100 & 0.10 & 0.15 & 0.75 \\
\bottomrule
\end{tabular}
\caption{Filtering candidate failures on the enriched human-adjudicated candidate-failure queue. Prec. is human-adjudicated precision, Quest. is the questionable rate, and FP is the false-positive rate; these are descriptive queue-level summaries.}
\label{tab:filtering}
\end{table}

\begin{table}[h]
\centering
\small
\setlength{\tabcolsep}{2.8pt}
\begin{tabular}{lrrrr}
\toprule
Model & C-Acc. & Q-Acc. & C-Susp. & Q-Susp. \\
\midrule
Qwen & 0.67 & 0.61 & 0.039 & 0.036 \\
OpenBio & 0.65 & 0.65 & 0.038 & 0.034 \\
MedGemma & 0.61 & 0.59 & 0.032 & 0.042 \\
BioMistral & 0.48 & 0.42 & 0.045 & 0.036 \\
Med42 & 0.69 & 0.70 & 0.030 & 0.030 \\
\bottomrule
\end{tabular}
\caption{Prompt and option-order robustness by model on 100 DDXPlus cases. C and Q denote clinical-summary and questionnaire prompts. Acc. and Susp. report the mean accuracy and strict suspicious-strength ratio over three candidate orders.}
\label{tab:prompt-order}
\end{table}

\begin{table}[h]
\centering
\small
\setlength{\tabcolsep}{3.0pt}
\begin{tabular}{lrrrr}
\toprule
Model & Drop & Neut. & Drop str. & Neut. str. \\
\midrule
Qwen & 2.38 & 2.17 & 0.035 & 0.029 \\
OpenBio & 0.61 & 0.14 & 0.024 & 0.012 \\
MedGemma & 1.60 & 0.86 & 0.030 & 0.025 \\
BioMistral & 1.38 & 0.15 & 0.039 & 0.020 \\
Med42 & 1.38 & 0.87 & 0.027 & 0.018 \\
\bottomrule
\end{tabular}
\caption{Perturbation-semantics check on 200 DDXPlus cases. Drop and Neut. report suspicious interactions per case under deletion and neutralization; str. reports the strict suspicious-strength ratio.}
\label{tab:neutralize}
\end{table}

Together, these checks isolate the stable, clinically reviewed subset behind stronger failure claims.

\section{Case Studies}
\label{sec:case-studies}

\begin{figure*}[t]
\centering
\includegraphics[width=0.98\textwidth]{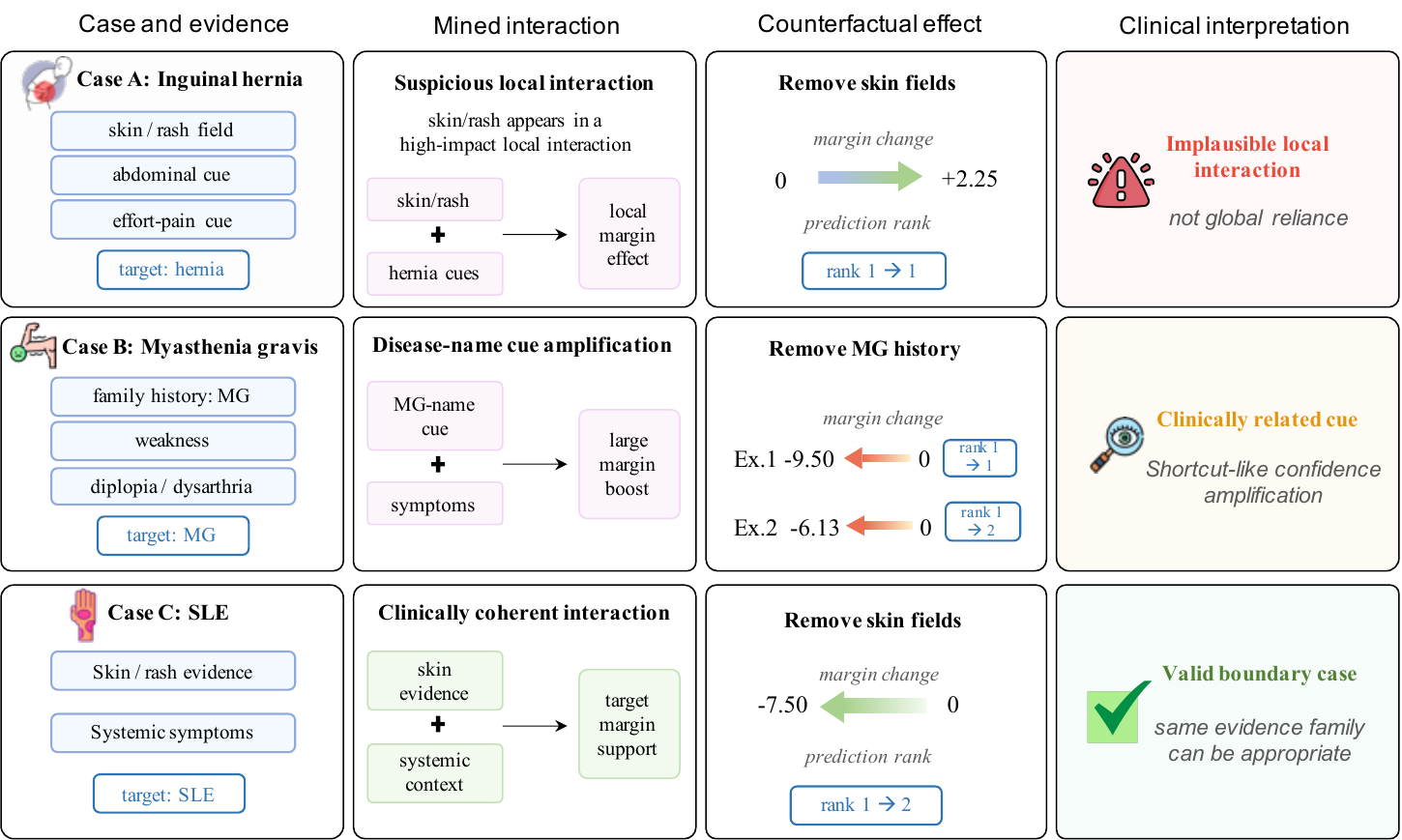}
\caption{Case-study summary. Each row contrasts the evidence pattern, mined interaction, targeted counterfactual effect, and clinical interpretation. The examples separate three outcomes: an implausible local interaction without global reliance, shortcut-like confidence amplification from a clinically related cue, and a valid boundary case where the same evidence family is clinically appropriate.}
\label{fig:case-studies}
\end{figure*}

Figure~\ref{fig:case-studies} summarizes three representative cases where interaction magnitude alone is insufficient: an implausible local interaction without global reliance, shortcut-like confidence amplification, and a valid diagnosis-relative boundary case. Additional details are in Appendix~\ref{app:additional-cases}.

\paragraph{Case A: implausible local interaction without global reliance.}
For inguinal hernia, skin or rash fields appear in high-impact interactions with abdominal and effort-pain cues. Clinically, these fields do not provide direct support for hernia. However, removing the skin fields increases the target margin and leaves the prediction rank unchanged. The supported claim is therefore local evidence-use incoherence, not global reliance on rash evidence.

\paragraph{Case B: clinically related but shortcut-like evidence.}
For myasthenia gravis, family history is clinically relevant, but the exact disease-name cue can dominate the diagnostic margin. Removing the MG-history phrase substantially lowers the target margin, and in one representative pair changes the top-ranked diagnosis. This supports a confidence-amplification claim rather than a simple irrelevant-feature claim.

\paragraph{Case C: a boundary case that protects against over-labeling.}
For systemic lupus erythematosus (SLE), skin evidence is clinically meaningful in the systemic case context. Removing skin fields lowers the target margin and can change the top-ranked diagnosis. This boundary case shows why evidence roles must be diagnosis-relative: the same evidence family can be implausible for hernia but appropriate for SLE.

Together, the cases show why the audit reports the clinical role of a score change, not only its magnitude. Case A shows local incoherence without global reliance. Case B supports a stronger shortcut-like confidence claim: removing the disease-name history cue sharply lowers the margin and can change the top diagnosis. Case C is the boundary condition: the same evidence family is appropriate when the audited diagnosis supplies the clinical bridge. These distinctions make the cases cross-checks rather than isolated anecdotes.

\section{Discussion and Conclusion}

These results suggest that an evidence-use audit should be reported as a structured profile, not as a binary list of suspicious features. Each high-effect interaction should be interpreted with the audited diagnosis, evidence role, margin direction, affected target or competitor, and stability under prompt, option-order, and perturbation changes. This is what separates patterns that can look similar numerically: clinically coherent differential conflict, local but implausible evidence use, and shortcut-like cues that amplify confidence. The model-wise and selection-sensitivity checks also show why this profile is useful: faithful or plausible support and differential conflict dominate across settings, while stable suspicious mechanisms form a smaller subset that needs clinical review.

Diagnostic accuracy alone cannot show whether medical LLMs use evidence appropriately. We introduced a behavioral audit that traces how controlled evidence interventions change diagnostic margins, interprets interactions with diagnosis-relative roles, and reserves failure claims for stable clinically reviewed mechanisms. Across three diagnostic datasets and five open-weight models, most interaction strength reflects plausible support or differential conflict, while the DDXPlus-enriched review sample identifies stable risks around absent or clinically local evidence. These findings motivate reporting evidence-use structure alongside diagnostic accuracy when evaluating medical LLMs, especially when evidence roles shift across candidates in real differential diagnosis.

\clearpage
\section*{Limitations}

This audit measures prompt-conditioned preferences under evidence interventions, not latent reasoning or clinical safety; interaction scores are behavioral signals, not internal causal explanations.

Results depend on evidence units, candidate sets, and role annotations \citep{bender2018data,gebru2021datasheets,mitchell2019model}. DDXPlus supports structured roles; CupCase and MedCase use extracted narrative units and coarser labels. Clinical review is enriched for high-strength and candidate-failure items, so precision, mechanism shares, model comparisons, and filtering values are descriptive rather than population estimates. We report Wilson and bootstrap intervals for selected review and mechanism summaries, but broader prevalence estimates would require probability sampling and more models. The order-ablation is a DDXPlus retrieval check, not proof that order-3 suffices generally.

Experiments cover five open-weight models, selected units, and multiple-choice option scoring. Option-letter log probabilities may reflect token or position bias; margins and order permutations reduce but do not calibrate this. Robustness checks cannot remove all prompt or candidate-set sensitivity, so failure claims are strongest for stable, clinically reviewed patterns.

\bibliography{ref}

\clearpage
\appendix

\section{Related Work}
\label{app:related-work}

\subsection{Medical Diagnostic Evaluation}

Medical LLMs are commonly evaluated with diagnostic benchmarks, clinical multiple-choice exams, or case-based question answering \citep{jin2019pubmedqa,jin2021disease,pal2022medmcqa,kung2023performance,singhal2023large,nori2023capabilities}. These evaluations provide an important first-order signal: whether a model can select the correct diagnosis or answer when given the full case. General-purpose benchmark suites and holistic evaluations make a similar tradeoff, aggregating broad task performance while leaving more detailed behavioral structure less visible \citep{hendrycks2020measuring,srivastava2023beyond,liang2022holistic,lin2022truthfulqa}. A correct final answer can arise from faithful symptom integration, memorized disease-name cues, over-weighted background history, or irrelevant evidence that happens to move the score in the right direction. Our audit is designed to complement diagnostic accuracy by measuring how diagnostic preferences change under controlled evidence interventions.

This distinction is especially important for diagnosis because clinical evidence is heterogeneous. Symptoms, signs, risk factors, negative findings, family history, and laboratory results do not play the same role. Clinical NLP resources such as MIMIC-III and emrQA helped make structured clinical records and record-grounded question answering available for model evaluation \citep{johnson2016mimic,pampari2018emrqa}. DDXPlus, CupCase, and MedCase expose the diagnostic version of this issue in different forms, ranging from structured differential-diagnosis fields to narrative case presentations \citep{fansi2022ddxplus,perets2025cupcase,wu2025medcasereasoning}. This motivates an evaluation framework that does not treat evidence as globally relevant or irrelevant, but instead asks whether the model's diagnostic preference changes in a clinically coherent way.

\subsection{Explanation, Attribution, and Evidence Interactions}

A large body of work studies explanations and attributions for language models, including local explanations, anchors, influence functions, gradient-based attribution, concept activation vectors, Shapley-style feature attribution, and rationale benchmarks \citep{ribeiro2016should,ribeiro2018anchors,koh2017understanding,sundararajan2017axiomatic,kim2018interpretability,lundberg2017unified,deyoung2020eraser}. These methods are useful for identifying influential tokens, rationales, or input regions, but many diagnostic decisions depend on combinations of findings rather than isolated evidence units. For example, fever, cough, wheezing, and prior asthma can each be weak or ambiguous alone, while their joint effect may shift the differential diagnosis substantially. This motivates our use of interaction effects over evidence subsets rather than only single-evidence attribution.

Our goal differs from recovering a model's hidden reasoning process. Work on explanation faithfulness shows why generated rationales, attention weights, or saliency maps alone should not be treated as direct evidence of model reasoning \citep{jain2019attention,serrano2019attention,adebayo2018sanity,jacovi2020towards,turpin2023language}. We instead audit observable diagnostic preferences: the model is repeatedly scored under controlled evidence subsets, and we analyze the resulting margin changes. The interaction score is therefore not a claim about internal causal computation. It is a structured way to expose which evidence combinations have large observable effects on the model's diagnostic preference, in the spirit of interaction indices that quantify joint feature effects \citep{grabisch1999axiomatic,sundararajan2020shapley,janizek2021explaining,tsang2017detecting}.

\subsection{Counterfactual Auditing and Regulated Domains}

Counterfactual and perturbation-based evaluations test whether model behavior changes when input evidence is removed, replaced, or otherwise controlled \citep{kaushik2019learning,gardner2020evaluating,ribeiro2020beyond}. Such tests are natural for regulated domains because they can expose reliance on inappropriate cues, complementing broader documentation and audit practices for deployed models \citep{bender2018data,gebru2021datasheets,mitchell2019model}. Safety and fairness work in language technology and health further motivates checking not only performance, but also the evidence and proxies that drive model decisions \citep{bender2021dangers,obermeyer2019dissecting}. However, medical diagnosis differs from settings where a feature can often be marked as globally forbidden or irrelevant. A negative interaction may reflect plausible differential diagnosis, and a background cue may be clinically meaningful in one case but overused in another.

Our framework therefore separates discovery from error assignment. Interaction mining proposes candidate mechanisms; clinical roles, robustness checks, and human validation determine whether those mechanisms support a failure claim. This differs from audits that treat any sensitivity to an undesirable feature as direct evidence of unreliability. In medicine, the stronger claim must be narrower: stable failures are those where the evidence role, interaction sign, case context, and counterfactual behavior remain clinically incoherent after review.

\section{Experimental Setting Details}
\label{app:experimental-setting}

\subsection{Datasets and Case Selection}

We use DDXPlus as the main controlled dataset and CupCase and MedCase as external case-style checks. DDXPlus provides structured patient findings and a differential diagnosis list, which makes it suitable for evidence-role analysis. CupCase provides case presentations with one correct diagnosis and distractor diagnoses. MedCase contains longer case prompts and diagnostic reasoning; we use it to test whether the broad interaction patterns persist in less structured clinical narratives.

For DDXPlus, we keep cases with a gold diagnosis, the provided differential list, at least two candidate diagnoses, and enough structured findings to form the audited evidence units. For CupCase and MedCase, we keep cases with a gold diagnosis, at least three usable candidate diagnoses, and enough patient information to form the required number of evidence units. DDXPlus contributes 500 validation cases with eight audited evidence units per case. CupCase and MedCase contribute 200 test cases each with six audited evidence units per case. The smaller number of units for the narrative datasets keeps the subset intervention space comparable while preserving clinically readable evidence units.

\begin{table}[h]
\centering
\small
\setlength{\tabcolsep}{4pt}
\begin{tabular}{llrp{0.36\columnwidth}}
\toprule
Dataset & Split & Units & Candidate source \\
\midrule
DDXPlus & validation & 8 & differential list \\
CupCase & test & 6 & answer + distractors \\
MedCase & test & 6 & answer + reasoning diagnoses \\
\bottomrule
\end{tabular}
\caption{Dataset construction used in the audit. ``Units'' is the number of audited evidence units selected per case.}
\label{tab:appendix-data}
\end{table}

\subsection{Evidence Unit Construction}

For DDXPlus, evidence units are built from structured findings. Each unit stores the underlying finding, its observed value, whether it is present or absent, and whether the finding is listed as a symptom or antecedent for the target diagnosis. We then assign diagnosis-relative roles by comparing the unit not only to the gold diagnosis but also to the other candidate diagnoses in the same case. This allows a unit to be marked as target-supporting, competitor-supporting, negated or absent, background/history, low-specificity, or other informative evidence.

When a case contains more findings than we can exhaustively audit, we select a balanced subset rather than simply taking the first fields. The selection prioritizes target-specific symptoms, target antecedents, competitor-only evidence, explicitly absent findings, low-specificity fields, and other informative units. This keeps the audit focused on evidence that can affect differential diagnosis while still including plausible failure triggers such as absent or weakly related fields.

For CupCase and MedCase, evidence units are extracted as short clinically meaningful sentences from the case presentation. We avoid using sentences that directly reveal the final diagnosis. The selected sentences are then treated as case evidence units and enumerated under the same subset-intervention procedure as DDXPlus. These narrative datasets do not support the same fine-grained DDXPlus role labels, so we use them mainly for broad mechanism analysis rather than detailed failure taxonomy.

\subsection{Candidate Diagnoses and Option Scoring}

For DDXPlus, candidate diagnoses come from the provided differential diagnosis list; if the gold diagnosis is not already present, it is added to the candidate set. For CupCase, the candidate set consists of the correct diagnosis and the provided distractors. For MedCase, the candidate set consists of the final diagnosis and additional diagnoses mentioned in the diagnostic reasoning, after de-duplication and length filtering.

All datasets are scored with the same multiple-choice format. Candidate diagnoses are mapped to option letters, and the model is asked to answer with only the option letter. We use the option-letter scores for all candidates from the same prompt, rather than relying only on the generated answer. The main experiments use a fixed candidate order; robustness checks repeat the scoring under alternative candidate orders and compare prediction agreement, margin correlation, and top-interaction overlap. In these checks, diagnoses are remapped to different letters and positions while the candidate set is held fixed, so changes reflect prompt and option-format sensitivity rather than a changed differential diagnosis.

\subsection{Models and Scoring Protocol}

The main experiments use Qwen2.5-7B-Instruct, OpenBioLLM-8B, MedGemma-4B, BioMistral-7B, and Llama3-Med42-8B. All models are evaluated with the same multiple-choice scoring protocol: each candidate diagnosis is assigned an option letter, the model is prompted to answer with only the letter, and candidate preferences are computed from next-token option-letter log probabilities. Scoring uses vLLM with temperature 0, one generated token, and max-logprobs 100; option-letter token IDs are validated with the prompt tokenizer. We do not use free-form rationales or generated explanations in the audit. Because option-letter scores can contain token-prior and position effects, all main interaction values use within-prompt margins and robustness checks test option-order permutations.

\subsection{Human and Clinical Review Protocol}

Clinical validation used high-strength interactions sampled across the main mechanism buckets. The review was performed by five senior PhD-trained medical reviewers. Each reviewer independently annotated every reviewed item under a blinded protocol. Reviewers saw the audited diagnosis, candidate diagnoses, evidence units in the interaction, interaction sign and strength, and the surrounding case context, but not the source model or provisional automatic mechanism bucket. They assigned one of three judgments: valid mechanism, questionable, or invalid/shortcut-like. The review is a clinical-coherence judgment about the evidence mechanism, not a patient-care judgment.

The validation set contains 130 reviewed interactions. Sampling across buckets prevents the review from being dominated by the most frequent mechanism and lets us test whether provisional failure labels have higher invalid rates than ordinary faithful-support or conflict/cancellation mechanisms. We computed inter-rater agreement on the independent first-pass labels before adjudication. All five reviewers gave the same label on 77\% of items, and agreement was high (Fleiss' $\kappa=0.82$). Disagreements were resolved by adjudication after independent review. The reported labels are adjudicated labels; questionable cases are reported separately.

\begin{table}[h]
\centering
\small
\setlength{\tabcolsep}{4pt}
\begin{tabular}{lp{0.54\columnwidth}}
\toprule
Property & Setting \\
\midrule
Reviewers & Five senior PhD-trained medical reviewers \\
Items & 130 high-strength interactions \\
Blinding & Source model and provisional bucket hidden \\
Independent labels & All five reviewers annotated each item \\
Agreement & Fleiss' $\kappa=0.82$; 77\% all-reviewer exact agreement \\
Resolution & Adjudication after independent review \\
\bottomrule
\end{tabular}
\caption{Clinical review protocol summary.}
\label{tab:appendix-review-protocol}
\end{table}

\section{Prompt Templates}
\label{app:prompts}

\subsection{Prompt Conventions}

All diagnosis prompts present evidence units as short case facts and candidate diagnoses as option-letter choices. The model is instructed to return only the option letter. The templates below show the representative prompt forms used for full-evidence scoring, subset interventions, neutralized interventions, and clinical review. Dataset-specific fields such as age, sex, and finding text are inserted into the same structure.

\begin{diagnosisprompt}{Full-evidence diagnosis prompt}
\small
You are given a structured clinical intake summary. Use only the primary patient findings below. Ignore administrative labels, prior assessments, source-file diagnoses, or information not specific to this patient.

Patient: [AGE]-year-old [SEX].

Clinical findings:

- [EVIDENCE UNIT 1]

- [EVIDENCE UNIT 2]

- [EVIDENCE UNIT 3]

Options:

A. [DIAGNOSIS A]

B. [DIAGNOSIS B]

C. [DIAGNOSIS C]

Answer with only the option letter.
\end{diagnosisprompt}

\begin{subsetprompt}{Subset intervention prompt}
\small
You are given a structured clinical intake summary. Use only the primary patient findings below. Some evidence from the original case is not shown.

Patient: [AGE]-year-old [SEX].

Visible patient information:

[EVIDENCE UNIT 1]

[EVIDENCE UNIT 2]

[EVIDENCE UNIT 3]

Options:

A. [DIAGNOSIS A]

B. [DIAGNOSIS B]

C. [DIAGNOSIS C]

Answer with only the option letter.
\end{subsetprompt}

\begin{neutralprompt}{Neutralized intervention prompt}
\small
You are given a structured clinical intake summary. Fields marked as not specified should not be assumed present or absent.

Patient: [AGE]-year-old [SEX].

Clinical findings:

[VISIBLE EVIDENCE UNIT]

[HIDDEN FIELD]: not specified.

Options:

A. [DIAGNOSIS A]

B. [DIAGNOSIS B]

C. [DIAGNOSIS C]

Answer with only the option letter.
\end{neutralprompt}

\begin{reviewprompt}{Clinical review form}
\small
Diagnosis under audit: [DIAGNOSIS]

Candidate diagnoses: [CANDIDATE LIST]

Evidence units in interaction: [EVIDENCE UNITS]

Interaction sign and strength: [SIGN], [STRENGTH]

Case context: [SHORT CASE CONTEXT]

Review judgment: valid mechanism / questionable / invalid or shortcut-like

Brief rationale: [FREE TEXT]
\end{reviewprompt}

\subsection{Prompt Variants for Robustness}

Prompt robustness variants preserve the same evidence and candidate list while changing wording around the task instruction, evidence presentation, and answer format. We use two main wording families. The clinical-summary variant presents evidence as a compact intake note. The questionnaire variant presents the same findings as question-answer pairs. Candidate-order robustness is tested separately by permuting the option order while keeping the same candidate set. These variants are used only for sensitivity analysis; the main reported scores use the base clinical-summary prompt.

\section{Evidence Roles and Audit Labels}
\label{app:roles-labels}

\subsection{Diagnosis-Relative Evidence Roles}

Evidence roles are assigned relative to an audited diagnosis. A unit can support the target diagnosis, support a competitor, exclude or weaken the diagnosis, modify prior plausibility as background or risk evidence, or be likely irrelevant in the local clinical context. These roles are directional expectations rather than hard rules, and they are used to interpret interaction signs rather than to assign errors automatically.

\begin{table*}[t]
\centering
\small
\setlength{\tabcolsep}{4pt}
\begin{tabular}{p{0.18\textwidth}p{0.31\textwidth}p{0.42\textwidth}}
\toprule
Role family & Directional expectation & Important caveat \\
\midrule
Supporting evidence & Should usually increase the margin for the audited diagnosis. & Support can be redundant with other findings, so the joint effect need not be additive. \\
Competitor-supporting evidence & Can reasonably lower the audited diagnosis margin by increasing a plausible alternative. & A negative interaction is not automatically a failure; it may reflect normal differential diagnosis. \\
Excluding or negated evidence & Should usually weaken diagnoses that require the absent finding. & Absence is often weak evidence, not proof against a diagnosis. Deletion means unavailable, not present. \\
Background or risk evidence & Can change prior plausibility or recurrence risk. & Background evidence can be valid in one case and overused in another. \\
Likely irrelevant or local evidence & Should have little direct effect without a clinical bridge to the diagnosis. & The same field family can be relevant for another diagnosis, such as skin findings for systemic lupus erythematosus. \\
\bottomrule
\end{tabular}
\caption{Diagnosis-relative evidence roles used to interpret mined interactions. Roles are directional expectations rather than automatic error labels.}
\label{tab:appendix-roles}
\end{table*}

\subsection{Automatic Mechanism Labels}

Automatic labels are provisional. Faithful target evidence, conflict/cancellation, redundancy, and background/history influence describe candidate mechanisms that may be clinically valid. Negated or absent evidence misuse and irrelevant evidence misuse are candidate failure mechanisms, but they require stability checks and clinical validation before being reported as adjudicated invalid mechanisms.

\begin{table*}[t]
\centering
\small
\setlength{\tabcolsep}{4pt}
\begin{tabular}{p{0.22\textwidth}p{0.34\textwidth}p{0.35\textwidth}}
\toprule
Mechanism label & Screening signal & How it is used \\
\midrule
Faithful target evidence & Target-supporting roles increase the target margin. & Treated as a plausible mechanism unless clinical review finds otherwise. \\
Differential conflict/cancellation & Evidence supports both target and competitors, or reduces the target margin while supporting an alternative. & Not treated as a failure by default. Used to characterize differential-diagnosis behavior. \\
Redundancy & Multiple related findings have a weaker joint effect than an additive view would suggest. & Usually plausible; helps avoid over-interpreting small or canceling effects. \\
Background/history influence & History, exposure, or risk evidence has a large effect on the margin. & Reviewed for possible overuse, but often clinically valid. \\
Negated or absent evidence misuse & An explicitly absent or excluding finding increases support in an unexpected direction. & Candidate failure; requires stability and clinical confirmation. \\
Irrelevant evidence misuse & A clinically local or unrelated field strongly affects the audited diagnosis. & Candidate failure; counterfactual validation checks whether the effect is local or changes the final preference. \\
\bottomrule
\end{tabular}
\caption{Provisional mechanism labels. Automatic labels retrieve candidate mechanisms; they are not final clinical judgments.}
\label{tab:appendix-labels}
\end{table*}

\subsection{Boundary Cases}

Some evidence families are only interpretable with diagnosis context. Skin evidence is suspicious for inguinal hernia, ambiguous for influenza, and clinically meaningful for systemic lupus erythematosus. Family history of myasthenia gravis is relevant, but exact disease-name mentions can still produce shortcut-like confidence amplification. These boundary cases are why the audit avoids global evidence labels and reports questionable cases separately from adjudicated invalid mechanisms.

\section{Additional Experimental Results}
\label{app:additional-results}

\subsection{Model-Wise Mechanism Distributions}

Table~\ref{tab:appendix-model-mechanisms} reports DDXPlus mechanism strength by model. The main pattern is consistent across models: faithful target evidence and conflict/cancellation account for most interaction strength, while negated/absent misuse is much smaller but clinically important. This supports the paper's main interpretation. The audit is not finding a large mass of obvious failures; it is finding a broad structure of plausible differential-diagnosis interactions plus a smaller set of stable failure candidates.

\begin{table}[h]
\centering
\small
\setlength{\tabcolsep}{3.0pt}
\begin{tabular}{lrrrr}
\toprule
Model & Faith. & Conflict & Back. & Neg. \\
\midrule
Qwen & 40.9 & 48.7 & 7.7 & 2.7 \\
OpenBio & 42.2 & 44.4 & 7.6 & 1.2 \\
MedGemma & 43.9 & 45.3 & 8.1 & 2.2 \\
BioMistral & 41.4 & 45.9 & 7.1 & 3.0 \\
Med42 & 40.9 & 48.6 & 8.4 & 1.7 \\
\bottomrule
\end{tabular}
\caption{DDXPlus mechanism strength by model (\%). Faith. denotes faithful or plausible support, Conflict denotes differential conflict/cancellation, Back. denotes background/history influence, and Neg. denotes negated/absent evidence misuse.}
\label{tab:appendix-model-mechanisms}
\end{table}

\subsection{Dataset-Wise Mechanism Differences}

Table~\ref{tab:appendix-dataset-mechanisms} reports mechanism strength by dataset. DDXPlus has a visible background/history component because its structured fields expose antecedents and risk factors. CupCase and MedCase are dominated by faithful/plausible support and conflict/cancellation because their evidence units are narrative clinical sentences and do not support the same fine-grained role decomposition. The important cross-dataset result is that conflict/cancellation remains large in all three datasets, which argues against treating negative interactions as automatic errors.

\begin{table}[h]
\centering
\small
\setlength{\tabcolsep}{3.0pt}
\begin{tabular}{lrrrr}
\toprule
Dataset & Faith. & Conflict & Back. & Neg. \\
\midrule
DDXPlus & 41.9 & 47.1 & 7.9 & 2.2 \\
CupCase & 54.1 & 45.8 & -- & -- \\
MedCase & 47.8 & 52.2 & -- & -- \\
\bottomrule
\end{tabular}
\caption{Mechanism strength by dataset (\%). Narrative datasets are mapped to broad faithful/plausible support and conflict/cancellation categories.}
\label{tab:appendix-dataset-mechanisms}
\end{table}

\subsection{Evidence-Unit Selection Sensitivity}

The main DDXPlus experiments use a balanced evidence-unit selector that prioritizes target-specific symptoms, target antecedents, competitor-only evidence, explicitly absent findings, low-specificity fields, and other informative units. To test whether the mechanism distribution depends on this selector, we compare it with a target-salient selector and a random-balanced selector averaged over three seeds. The target-salient selector prioritizes target-supporting symptoms and antecedents, while the random-balanced selector samples from the same role families without fixed priority.

\begin{table}[h]
\centering
\small
\setlength{\tabcolsep}{3.0pt}
\begin{tabular}{lrrrrr}
\toprule
Selection & Faith. & Conflict & Susp. & Neg. & Overlap \\
\midrule
Main balanced & 41.9 & 47.1 & 2.2 & 2.2 & 1.00 \\
Target-salient & 47.0 & 47.0 & 2.4 & 2.1 & 0.49 \\
Random-balanced & 45.0 & 49.0 & 2.8 & 2.7 & 0.48 \\
\bottomrule
\end{tabular}
\caption{Sensitivity to DDXPlus evidence-unit selection. Faith. and Conflict report mechanism-strength percentages. Susp. denotes candidate suspicious-mechanism strength, Neg. denotes negated/absent misuse strength, and Overlap is top-interaction overlap with the main balanced selector. Random-balanced reports the mean over three seeds.}
\label{tab:unit-selection-sensitivity}
\end{table}

Across the alternative selectors, faithful/plausible support and differential conflict/cancellation remain the dominant mechanisms. Suspicious and negated/absent misuse strength remains small but does not disappear. The moderate top-interaction overlap is expected because the audited evidence units change; the stability claim is therefore about mechanism-level structure rather than exact-interaction invariance.

\subsection{Interaction Order and Threshold Sensitivity}

Table~\ref{tab:appendix-order} reports the DDXPlus interaction-order distribution. Third-order interactions account for the largest share of count and strength in the selected high-strength set, but first- and second-order interactions remain substantial. This is useful for interpretation: single-evidence effects alone miss much of the observable diagnostic preference structure, while low-order interactions are still small enough to inspect clinically.

\begin{table}[h]
\centering
\small
\begin{tabular}{lrrr}
\toprule
Order & Count & Strength & Mean \\
\midrule
1 & 14.3 & 18.0 & 2.98 \\
2 & 26.2 & 27.5 & 2.48 \\
3 & 59.5 & 54.4 & 2.16 \\
\bottomrule
\end{tabular}
\caption{DDXPlus interaction distribution by order. Count and strength are percentages; Mean is average absolute interaction effect.}
\label{tab:appendix-order}
\end{table}

We use high-strength filtering to make clinical review feasible. The threshold is therefore an audit design choice rather than a claim that lower-strength interactions are unimportant. In practice, the main qualitative conclusions are stable under reasonable filtering: faithful/plausible support and conflict/cancellation dominate the interaction mass, while DDXPlus-adjudicated invalid mechanisms concentrate in a smaller set of absent or clinically unrelated evidence patterns.

\subsection{Accuracy Versus Evidence-Use Structure}

Accuracy and evidence-use structure measure different properties. In the five-model DDXPlus sample, the model ordering by final accuracy does not match the ordering by faithful target-evidence strength or conflict/cancellation strength. We report this only as a qualitative sample observation, not as an inferential estimate of a broader model population. Its role is to motivate the audit: a model can have relatively high final accuracy while still exhibiting many conflict/cancellation interactions or a nontrivial number of candidate evidence-use failures.

\section{Clinical Validation Details}
\label{app:clinical-validation-details}

\subsection{Sampling Protocol}

The reviewed set is sampled from high-strength interactions across mechanism buckets. Sampling across buckets prevents the validation from being dominated by the most frequent mechanism and allows the review to test whether provisional failure labels have higher invalid rates than control mechanisms.

\begin{table}[h]
\centering
\small
\begin{tabular}{lr}
\toprule
Review bucket & Reviewed \\
\midrule
Faithful target evidence & 30 \\
Conflict/cancellation & 40 \\
Background/history & 20 \\
Negated/absent misuse & 20 \\
Control/unclear & 20 \\
\midrule
Total & 130 \\
\bottomrule
\end{tabular}
\caption{Clinical validation sample by provisional mechanism bucket.}
\label{tab:appendix-review-sample}
\end{table}

The sample is intentionally enriched for high-strength interactions and candidate failure buckets. It should therefore be interpreted as a validation set for the audit pipeline, not as an estimate of the population-level frequency of every failure type.

\subsection{Review Rubric}

A valid mechanism is clinically plausible given the diagnosis and case context. A questionable mechanism is not clearly wrong but lacks enough support or depends on a debatable interpretation. An invalid or shortcut-like mechanism reflects an implausible evidence use, such as an absent finding increasing support in the wrong direction or a clinically local field affecting an unrelated diagnosis.

Reviewers are asked to judge the evidence mechanism rather than the full diagnostic quality of the case. For example, a case can have the correct final diagnosis but still contain an invalid local interaction, and a negative interaction can be valid if it reflects evidence that supports a competing diagnosis. This distinction is central to the audit: clinical review confirms whether an interaction is coherent, while accuracy only checks the final answer.

\subsection{Additional Review Outcomes}

The main text reports aggregate review outcomes. The additional confidence breakdown in Table~\ref{tab:appendix-review-confidence} shows that the adjudicated invalid bucket is not driven only by low-confidence judgments. The negated/absent misuse bucket has both valid and invalid examples, but the invalid examples are concentrated and clinically concrete: no-travel signals supporting unrelated diagnoses, and skin/rash/no-peeling fields affecting inguinal hernia or influenza.

\begin{table}[h]
\centering
\small
\setlength{\tabcolsep}{3.0pt}
\begin{tabular}{lrr}
\toprule
Bucket & High & Medium \\
\midrule
Faithful target & 21 & 9 \\
Conflict/cancel. & 28 & 12 \\
Background/history & 15 & 5 \\
Negated misuse & 10 & 10 \\
Control/unclear & 12 & 8 \\
\bottomrule
\end{tabular}
\caption{Reviewer confidence by bucket for the 130 reviewed interactions.}
\label{tab:appendix-review-confidence}
\end{table}

\section{Robustness and Sensitivity Details}
\label{app:robustness-details}

\subsection{Filtering Definitions}

The implementation enumerates interventions over the selected audited evidence units. On DDXPlus, non-audited case facts are held fixed in the prompt; on CupCase and MedCase, the extracted evidence units are the audited units. Unless otherwise stated, interactions are computed on the answer-margin value function, with maximum order $|S|\leq 3$. We call an interaction salient or high-strength when $|I_y(S)|\geq 0.5$. Aggregate strength ratios are computed over all salient interactions, while review tables use the highest-strength interactions retained for inspection.

The strict suspicious set is intentionally narrower than all non-faithful interactions. It contains only salient interactions whose automatic label begins with suspicious: positive interactions containing negated or absent evidence, positive interactions made only of low-specificity or other-informative units, or positive interactions made from competitor-only evidence. Potential history or risk overuse and canceling or redundant target interactions are tracked separately. The strict suspicious-strength ratio is the sum of $|I_y(S)|$ over strict suspicious interactions divided by the total salient interaction strength.

Robustness files are regenerated with the same scoring and interaction code under changed prompt wording, three candidate-order seeds, and either deletion or neutralization of hidden selected units. In the neutralized setting, a hidden selected unit is kept in the prompt with the value ``Not specified'' rather than removed. The filtering sequence in Table~\ref{tab:filtering} is evaluated on an enriched candidate-failure review queue using the same clinical-coherence rubric as the human validation protocol: confirmed invalid or shortcut-like, questionable, or false positive/clinically valid. Precision is the confirmed invalid or shortcut-like share among interactions that survive each filter. The random control set is sampled from high-strength interactions not prioritized by the suspicious-mechanism rules. Because the queue is intentionally enriched for candidate failures, the precision values are filter-validation quantities rather than estimates of population-level failure prevalence.

\subsection{Prompt Wording and Candidate Order}

Prompt wording checks measure prediction agreement across wording variants. Candidate-order checks measure whether option reordering changes margins and top interactions. Because the answer is scored through option letters, these checks also probe option-letter and option-position calibration artifacts. They are intended to detect prompt artifacts, not to prove that the model is invariant to all prompt changes.

Prediction agreement is the fraction of cases where the top diagnosis is unchanged across prompt variants. Margin correlation compares the vector of diagnosis margins before and after candidate-order changes. Top-interaction overlap compares the highest-strength mined interactions across variants. We report these measures together because they capture different failure modes: a model can preserve its final answer while changing margins, or preserve margins while changing which evidence interactions appear in the top-ranked set.

\subsection{Drop Versus Neutralized Perturbations}

Deletion makes evidence unavailable, while neutralization replaces hidden fields with a not-specified statement. The neutralized setting is a conservative check for whether suspicious interactions depend only on removing text from the prompt. This is particularly important for negated findings, where deleting an absent symptom should not imply that the symptom is present.

For each suspicious interaction family, we compare whether the interaction survives under neutralized perturbations. Survival means that the same evidence family remains a high-effect candidate mechanism after hidden fields are represented as unspecified rather than removed. We do not require exact numerical equality between deletion and neutralization, because the prompts are semantically different; instead, we use survival as a robustness screen before making stronger failure claims.

\subsection{Random Controls}

Random controls estimate how often the pipeline would produce clinically adjudicated invalid mechanisms from interactions not selected by the suspicious-mechanism filter. The control set is useful because it separates the precision of the audit from the base rate of questionable interactions in the data.

The random-control set is sampled from interactions that are not prioritized by the candidate failure rules. Its low adjudicated precision and high false-positive rate in Table~\ref{tab:filtering} support the filtering design: the stable suspicious subset is not merely a random collection of high-strength interactions. At the same time, the control set prevents overclaiming by showing that interaction strength alone is insufficient for clinical failure assignment.

\section{Additional Case Studies}
\label{app:additional-cases}

\subsection{Negated and Absent Evidence Misuse}

The clearest invalid review examples involve absent or negated fields that increase support in an unexpected direction. In several cases, ``no recent travel'' appears in positive interactions for allergic sinusitis, asthma exacerbation, or Guillain-Barre syndrome. The issue is not that travel history is never relevant in medicine; it is that the absence of recent travel has no clear positive role for these diagnoses in the displayed case context.

Other invalid examples involve absent or local skin findings. For inguinal hernia, fields such as rash color or whether lesions peel off enter positive interactions with abdominal or effort-pain evidence. Clinical review marks these as invalid because skin lesions and peeling behavior do not provide direct support for inguinal hernia. These examples motivate the polarity and specificity language used in the main paper: the failures are not generic sensitivity to text, but specific misuse of absent or clinically local evidence.

\subsection{Skin and Rash Evidence Transfer}

Skin and rash fields are useful boundary cases because their relevance is highly diagnosis-dependent. They are clinically meaningful for systemic lupus erythematosus and other dermatologic or inflammatory conditions, but they are suspicious when they influence unrelated diagnoses such as inguinal hernia. In influenza cases, the interpretation can be intermediate: removing skin fields can improve the target margin by suppressing dermatologic or allergic competitors, which means the field acted more like a distractor than simple positive evidence for influenza.

This pattern is why the audit uses both interaction mining and targeted counterfactual validation. A mined interaction can show that skin evidence affects the margin, but the counterfactual test clarifies whether the field supports the target, amplifies a competitor, or only changes the margin without changing the final rank.

\subsection{Plausible Differential Conflict}

Large negative or canceling interactions are common and often clinically valid. A respiratory symptom can support bronchitis, pneumonia, influenza, asthma exacerbation, or pulmonary embolism depending on the rest of the case. Similarly, two pieces of target evidence can appear redundant if they express overlapping clinical information. Clinical validation confirms this distinction: the conflict/cancellation bucket contains no adjudicated invalid cases in the reviewed sample.

These cases are important controls. They show that the audit does not equate mathematical interaction strength with model failure. The failure label is reserved for interactions where the evidence role and direction are clinically incoherent after robustness filtering and review.

\subsection{Counterfactual Case Details}

The targeted counterfactual validation focuses on representative suspicious patterns surfaced by the interaction audit. For no-travel cases, the counterfactuals remove the travel field, mark it as unspecified, or flip it to recent travel present. For skin/rash cases, the counterfactuals remove skin fields, neutralize them as unrelated or unspecified, or replace them with unrelated administrative text. The goal is to test whether a suspicious local interaction actually controls the final diagnostic preference.

The counterfactuals reveal two mechanisms. First, irrelevant evidence can directly increase the target margin in clinically implausible ways. Second, irrelevant evidence can amplify a competing diagnosis and thereby reduce the target margin. The second mechanism is especially important for hernia cases, where removing skin/rash fields can move a model away from dermatologic or allergic competitors and back toward inguinal hernia. This is why the paper avoids the simplistic claim that ``rash supports hernia'' and instead reports a broader irrelevant-field distortion mechanism.

\subsection{Expanded Stratified Counterfactuals}

After the representative case studies, we ran an expanded stratified counterfactual validation on 86 DDXPlus records and all five models. Each record was scored with a clean prompt and three counterfactual variants. The primary analysis uses the clean-correct subset, because evidence-use claims are most interpretable when the model initially selected the correct diagnosis.

\begin{table}[h]
\centering
\scriptsize
\setlength{\tabcolsep}{2.0pt}
\resizebox{\columnwidth}{!}{%
\begin{tabular}{lrrrrrrr}
\toprule
Stratum & Rec. & Strong & Rank & T$\downarrow$ & C$\uparrow$ & C$\downarrow$ & Stable \\
\midrule
High no-travel & 11 & 11.6 & 0.7 & 1.4 & 20.4 & 12.2 & 63.3 \\
Low no-travel & 20 & 5.6 & 1.1 & 0.4 & 20.7 & 6.7 & 69.3 \\
High skin-hernia & 14 & 29.3 & 4.6 & 5.2 & 8.0 & 44.3 & 33.9 \\
High skin-other & 12 & 32.1 & 3.8 & 3.8 & 19.2 & 46.2 & 24.4 \\
Plausible skin & 24 & 45.0 & 12.5 & 13.1 & 35.2 & 14.7 & 33.6 \\
Low skin & 5 & 27.8 & 4.2 & 4.2 & 38.9 & 0.0 & 55.6 \\
\bottomrule
\end{tabular}
}
\caption{Expanded stratified counterfactual validation on clean-correct model-case pairs. Strong means $\lvert\Delta \mathrm{score}_y\rvert\geq 0.5$, $\lvert\Delta v_y\rvert\geq 1.0$, or a target-rank change. T and C denote target and strongest competitor movements after the counterfactual edit. Values except Rec. are percentages.}
\label{tab:appendix-stratified-counterfactual}
\end{table}

The expanded analysis refines the main text. We count a strong change when the counterfactual edit changes the target score by at least 0.5 log-probability units, changes the target margin by at least 1.0, or changes the target diagnosis rank. No-recent-travel cases are only modestly more sensitive than their low-effect controls, so we treat them as a secondary polarity example. Skin/rash cases are stronger: the high skin-hernia and high skin-other strata show larger strong-change rates, often through competitor suppression rather than direct target-score movement. The plausible-skin control also changes substantially, which is the expected boundary case when skin evidence is clinically relevant.

\end{document}